\documentclass[11pt]{article}
\usepackage{hyperref}

\usepackage{amsfonts, gensymb}
\usepackage{graphicx}
\usepackage{bm}
\usepackage{booktabs}
\usepackage{array} 
\usepackage{times}
\usepackage{fullpage}
\usepackage{color, soul}
\usepackage{natbib}
\usepackage[margin=1in]{geometry}

\usepackage{authblk}  
\renewcommand{\so}[1]{\mathfrak{so}(#1)} 
\title{Zero-shot Sim-to-Real Transfer for Reinforcement Learning-based Visual Servoing of Soft Continuum Arms}

\author{
  Hsin-Jung Yang$^{1,*}$, Mahsa Khosravi$^{1,*}$, Benjamin Walt$^{2,*}$, Girish Krishnan$^{2}$, Soumik Sarkar$^{1,**}$\\
  $^1$Iowa State University, $^2$University of Illinois Urbana-Champaign \\\vspace{0.3cm}$^*$Equal contribution, $^{**}$Corresponding author\\
  \vspace{0.5cm} 
\texttt{\{hjy, mahsak, soumiks\}@iastate.edu, \{walt, gkrishna\}@illinois.edu}
}

\hypersetup{
    colorlinks=true,       
    linkcolor=blue,        
    citecolor=blue,       
    urlcolor=blue          
}
\date{}
\begin{document}

\maketitle

\begin{abstract}%
 Soft continuum arms (SCAs) soft and deformable nature  presents challenges in modeling and control due to their infinite degrees of freedom and non-linear behavior. This work introduces a reinforcement learning (RL)-based framework for visual servoing tasks on SCAs with zero-shot sim-to-real transfer capabilities, demonstrated on a single section pneumatic manipulator capable of bending and twisting. The framework decouples kinematics from mechanical properties using an RL kinematic controller for motion planning and a local controller for actuation refinement, leveraging minimal sensing with visual feedback. Trained entirely in simulation, the RL controller achieved a 99.8\% success rate. When deployed on hardware, it achieved a 67\% success rate in zero-shot sim-to-real transfer, demonstrating robustness and adaptability. This approach offers a scalable solution for SCAs in 3D visual servoing, with potential for further refinement and expanded applications.%
\end{abstract}


\newcommand{\reals}{\mathbb{R}}

\begin{figure}[h]
\centering
\includegraphics[width=.90\linewidth]{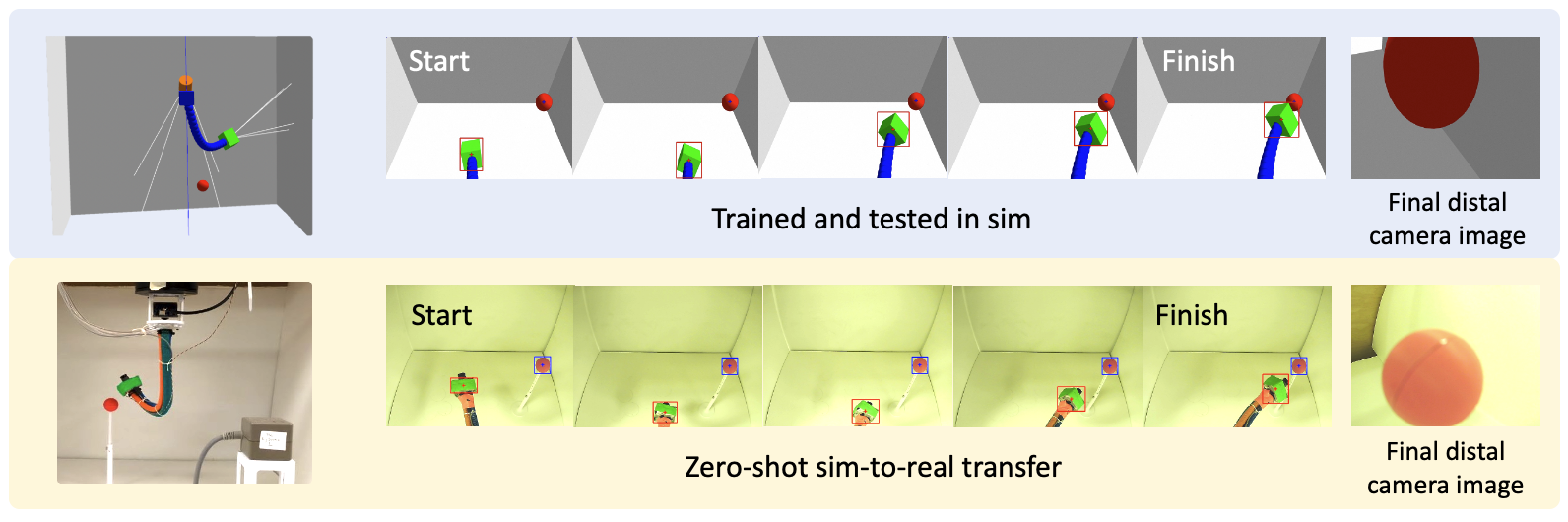}

\caption{{Overview of the proposed RL-based visual servoing control framework for SCAs with zero-shot sim-to-real transfer capability. The framework is used to visual servo to view a target as shown here, in sim (top) and on real hardware (bottom). The image sequence illustrates the base camera views after each policy step, with the final distal camera view (right) showcasing the RL-based controller’s ability to locate and center the target. Demo video in~\href{https://tinyurl.com/53f5vdje}{supplementary materials}. 
}}

\end{figure}

\section{Introduction}
Soft continuum arms (SCAs) are increasingly recognized for their ability to safely and effectively interact with complex, unstructured environments. Their ability to conform and apply gentle forces makes them ideal for tasks such as handling delicate objects or working in close proximity to humans~\citep{chen2022review, zongxing2020research, banerjee2018soft, chen2021soft, venter2017self}. However, their soft and deformable nature introduces challenges for modeling and control. Learning-enabled methods, such as model-free reinforcement learning (RL), offer a promising solution by learning behaviors directly from data rather than relying on analytically derived models~\citep{falotico2024learning}. 
Despite these advantages, one of the primary obstacles to deploying SCAs in real-world is the sim-to-real transfer, where policies trained in simulation fail to generalize well on physical systems. For SCAs, this challenge is amplified due to their unique physical characteristics. Unlike rigid robots, SCAs exhibit continuous deformation and high compliance, leading to non-linear behaviors that are difficult to model and generalize~\citep{rus2015design}. While prior work has demonstrated sim-to-real transfer for rigid robots using low-fidelity models and staged fine-tuning in simulation and on hardware~\citep{leguizamo2022deep}, such methods are not easily extended to SCAs due to their complex dynamics. Moreover, effective control of SCAs often requires significant sensing capabilities, such as accurate positional tracking or detailed environmental feedback, to account for their dynamic interactions with the environment. While high-fidelity simulations, such as the Cosserat rod model~\citep{till2019real, 9420666, xun2023cosseratrodbaseddynamicmodeling}, could theoretically address these discrepancies, they are computationally expensive and unsuitable for RL, which relies on large-scale data collection.

Existing RL-based approaches have attempted to bridge the sim-to-real gap but often remain constrained by sensing limitations or task-specific designs. For instance,~\citep{satheeshbabu2019open, wu2020position, morimoto2021model, li2024vision} implemented RL-based methods for 2D tasks with minimal sensing setups, such as single-camera systems or onboard sensors. However, these solutions struggled to extend to more high-dimensional scenarios. In contrast,~\citep{satheeshbabu2020continuous} extended RL to 3D navigation using Vicon motion capture systems, achieving high accuracy at the cost of extensive infrastructure.~\citep{thuruthel2018model} employed multi-sensor feedback for dynamic control, which showcases adaptability and precision but requires significant sensing resources. Meanwhile, some studies validated their methods without testing or deploying policies on physical hardware~\citep{goharimanesh2020fuzzy}. Notably, none of these works have demonstrated zero-shot sim-to-real transfer, leaving a critical gap in the field. These limitations often arise from the inherent challenges of scaling RL frameworks to address the non-linear dynamics of SCAs while maintaining robust sim-to-real transfer capabilities. A summary of relevant methods is provided in Table~\ref{tab:related_works}. Please see  the~\href{https://tinyurl.com/53f5vdje}{supplementary materials} for detailed related works.

To address these challenges, we propose a framework for SCAs that decouples kinematics from their mechanical properties by employing two components: an \textit{\textit{RL kinematic controller}} and a \textit{\textit{local controller}}. The \textit{RL kinematic controller} focuses on learning high-level kinematic policies, such as desired curvature and torsion, while the \textit{local controller} translates these commands into actuation signals, compensating for dynamic uncertainties and physical variations. By focusing on kinematics goals rather than full dynamic fidelity, our approach abstracts away complexities related to actuation and mechanical properties, accelerating the RL training process. Moreover, our framework leverages a minimum sensing approach, reducing the reliance on extensive setups such as multi-camera tracking systems, while primarily leveraging visual feedback from cameras and measurements from simple trackers. The above combination not only reduces computational burden, but also enhances transferability of learned policies by making them less sensitive to the underlying physical system.

We validate our framework using the BR2 manipulator~\citep{uppalapati2018design} through experiments demonstrating zero-shot sim-to-real transfer in visual servoing tasks in 3D space. Our results show that we achieved zero-shot sim-to-real transfer with a 67$\%$ success rate, showcasing the effectiveness of our proposed framework in bridging the sim-to-real gap. In summary, this paper introduces a control framework that decouples the kinematics of SCAs from their mechanical properties, simplifying policy learning and enhancing policy transferability. By leveraging minimum sensing, the approach negates the need for extensive sensing, while integrating a \textit{local controller} to handle real-world variations. We demonstrate the effectiveness of this framework through zero-shot sim-to-real transfer of an \textit{RL kinematic controller} on the BR2, achieving high accuracy in visual servoing tasks for 3D navigation and object tracking in both simulation and real-world experiments.
\begin{table}[h!]
\centering
\renewcommand{\arraystretch}{1.5} 
\resizebox{\textwidth}{!}{%
\begin{tabular}{>{\raggedright\arraybackslash}p{3.5cm}p{3cm}ccccc}
\toprule
\textbf{Literature} 
& \textbf{Goal} 
& \textbf{Model-Free}
& \textbf{Minimal Sensing} 
& \textbf{Closed-Loop} 
& \textbf{3D Task Space}
& \textbf{Zero-Shot Sim-to-Real Transfer} \\
\midrule
\cite{thuruthel2018model} 
& Dynamic control 
& $\times$
& $\times$ 
& \checkmark 
& \checkmark 
& $\times$ \\
\cite{satheeshbabu2019open} 
& Position control  
& \checkmark
& \checkmark 
& $\times$ 
& $\times$ 
& $\times$  \\
\cite{satheeshbabu2020continuous} 
& Path tracking 
& \checkmark
& $\times$ 
& \checkmark 
& \checkmark 
& $\times$ \\
\cite{morimoto2021model} 
& Pose control 
& \checkmark
& \checkmark 
& \checkmark 
& $\times$ 
& $\times$ \\
\cite{wu2020position} 
& Position control 
& $\times$
& $\times$ 
& \checkmark 
& $\times$ 
& $\times$ \\
\cite{goharimanesh2020fuzzy} 
& Trajectory tracking 
& $\times$
& $\times$ 
& \checkmark 
& $\times$ 
& $\times$ \\
\cite{li2024vision} 
& Visual servoing  
& \checkmark
& \checkmark 
& \checkmark 
& $\times$ 
& $\times$ \\
\textbf{Present Work} 
& Visual servoing
& \checkmark
& \checkmark
& \checkmark 
& \checkmark 
& \checkmark \\
\bottomrule
\end{tabular}%
}
\caption{A comparison of RL-based approaches for control of SCAs, highlighting key features. Minimal Sensing refers to methods that optimize sensory inputs relative to the complexity of the task, such as leveraging onboard sensors, or a limited number of cameras instead of extensive systems such as multi-camera tracking setups.}
\label{tab:related_works}

\end{table}

\section{Preliminaries}
\subsection{Modeling the BR2 with Constant Curvature and Constant Torsion Model}
The BR2 manipulator~\citep{uppalapati2018design} is a unique type of SCA with a parallel architecture composed of soft pneumatic actuators known as Fiber Reinforced Elastomeric Enclosures (FREEs). Unlike many existing soft manipulators, the BR2 utilizes an asymmetric configuration of FREEs, enabling it to bend and rotate simultaneously and achieve complex spatial deformation patterns. This design allows the BR2 to navigate around obstacles with enhanced flexibility while maintaining a parallel structural framework, which contributes to precise and adaptive control.

Modeling the BR2 requires accounting for both bending and torsional deformations. To achieve this, the BR2 is parameterized along its length, with the position $r(s)$ and orientation $R(s)$ defined at each point $s$ along its length. When actuated, the arm experiences a bending strain, $\kappa(s)$, and torsional strain, $\tau(s)$. These strains, when assuming negligible shear and stretching strains, can be related to the pose via the following differential equations: 
\(
r'\left(s\right) = R\left(s\right)\bm{v}\left(s\right), \quad R'\left(s\right) = R\left(s\right)\bm{\hat{u}}\left(s\right)
\)
where $\hat{\left[\,\cdot\,\right]}$ is the usual mapping from $\reals^{3}$ to $\so{3}$, $\bm{u} = [\kappa\left(s\right), 0 , \tau\left(s\right)]^{\top}$, and $\bm{v} = [0, 0, 1]^{\top}$. When combined with equations derived from static equilibrium, these differential equations form the Kirchhoff rod solution.  In this work, we assume $\kappa$ and $\tau$ are constant and thus form a constant curvature and torsion model. This simplified model provides a closed-form solution to relate pose and configuration to the strains ($\kappa$ and $\tau$), enabling rapid simulation modeling that is essential for the extensive training data required for effective RL. 


\subsection{Deep Reinforcement Learning}
Deep reinforcement learning (DRL)~\citep{mnih2013playing} extends traditional RL by leveraging the representational power of deep neural networks. It operates within the framework of a Markov Decision Process (MDP), characterized by a four-tuple $(S, A, P, R)$, where: $S$ is the state space, representing all possible states $s_t$ at time $t$; $A$ the action space, defining the set of all possible actions $a_t$ at time $t$; $P$ the transition probability function, which specifies the likelihood of transitioning from state $s_t$ to state $s_{t+1}$ given an action $a_t$; $R: S \times A \times S' \rightarrow \reals$ defines the reward function.

The agent’s decision-making process is governed by policy $\pi$, which maps the state $s_t$ to an action $a_t$. In DRL, this policy is parameterized by neural networks, represented as $\pi_{\theta}$, where $\theta$ denotes the learnable parameters of the neural network. The goal of DRL is to find an optimal policy $\pi_{\theta}^*$ that maximizes the expected reward over an episode $\tau$, defined as a sequence of states and actions $(s_0, a_0, s_1, a_1, \dots, s_T, a_T)$. The expected reward is expressed as $J(\theta) = \mathbb{E}_{\tau \sim \pi_{\theta}}\left[\sum_{t=0}^{T} \gamma^t R(s_t, a_t, s_{t+1})\right]$, where $\gamma$ is the discount factor and $T$ is the episode length.

DRL algorithms can be broadly classified into on- and off-policy approaches. Among these, Proximal Policy Optimization (PPO)~\citep{schulman2017proximal} and Soft Actor-Critic (SAC)~\citep{haarnoja2018soft} are two popular algorithms in their respective categories. On-policy algorithms like PPO optimizes policies by directly sampling data from the current policy, making them inherently less sample efficient. In contrast, off-policy algorithms like SAC utilize a replay buffer to leverage past experience, which significantly improves sample efficiency. SAC also incorporates entropy regularization, which encourages exploration by balancing the trade-off between expected rewards and policy stochasticity. This mechanism enhances robustness during training and reduces the likelihood of convergence to suboptimal policies. In this work, we adopt SAC due to its sample efficiency and entropy regularization, which together make it well-suited for our problem.


\section{Methodology}

Our goal is to develop an RL-based controller for visual servoing of SCAs in 3D space with zero-shot sim-to-real transfer. To enable generalizable control, we propose 1) a novel two-layer framework that decouples kinematics from mechanical properties and 2) adopt an open-vocabulary object detection model for feature extraction. An \textit{RL kinematic controller} plans high-level motion in configuration space, while a \textit{local controller} translates these plans into actuation, compensating for physical uncertainties. Visual feedback from a distal and a base camera is processed by the object detector to extract task-relevant features that guide the RL policy.


\subsection{Decoupling Kinematics and Mechanical Properties}
\begin{figure}[h!]
\centering
\includegraphics[width=0.9\linewidth]{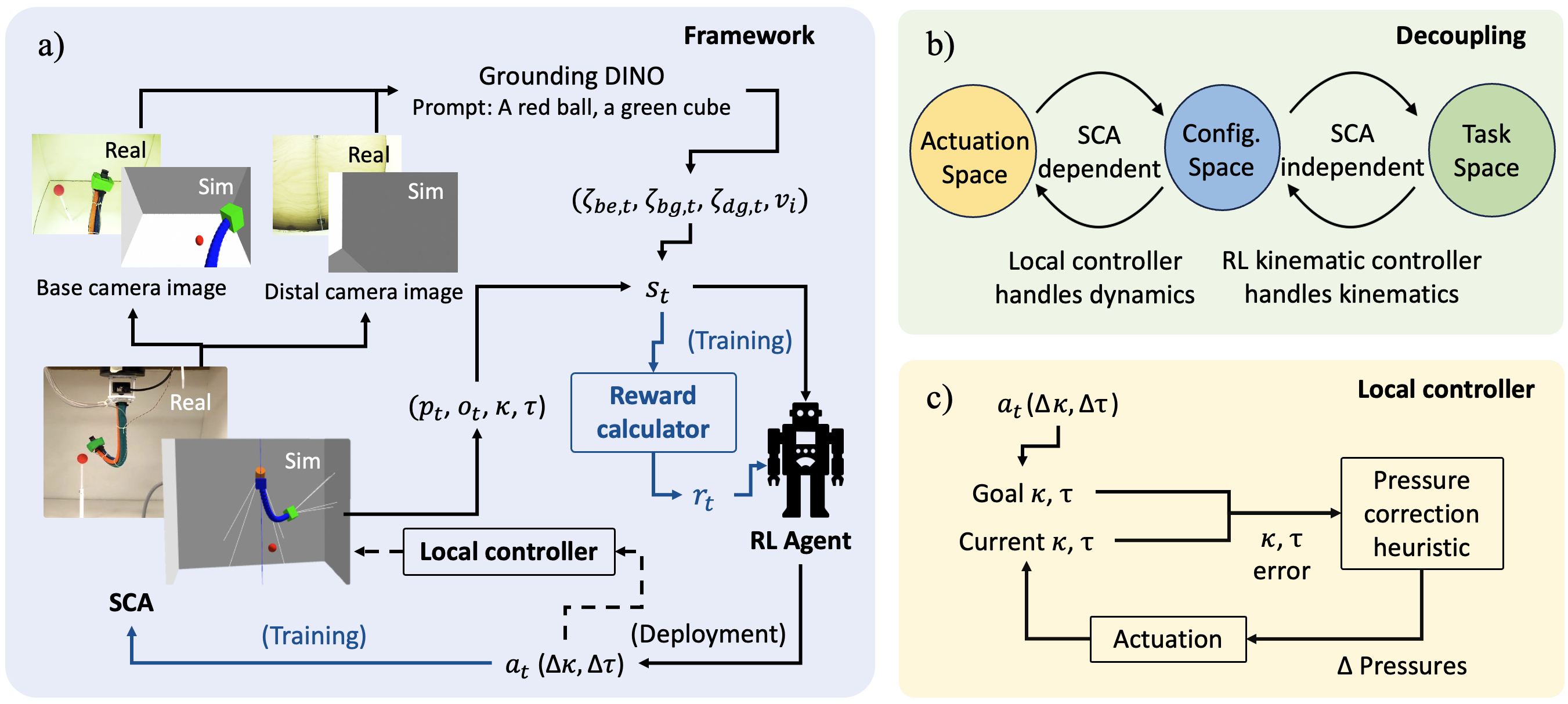}
\caption{\textbf{a)} Training and deployment framework of the RL kinematic controller. 
During training (black + blue paths), the RL agent learns a policy in simulation. In deployment (black + dashed paths), the trained policy outputs kinematic actions, translated into actuation by the \textit{local controller}. \textbf{b)} Decoupling kinematics and mechanical properties: The \textit{RL kinematic controller} handles the kinematics of the SCA, which is independent to specific hardware variations. The \textit{local controller} handles the dynamics of the SCA during operation, which tries to achieve the goal configuration determined by the \textit{RL kinematic controller}. \textbf{c)} The \textit{local controller} achieves the goal configuration without using a Configuration-to-Actuation map. Current $\kappa$,$\tau$ are estimated and the configuration error is passed to the heuristic, which generates an change in actuation.  This process is iterated until the goal configuration is achieved.} 

\label{fig:framework_and_controller}
\end{figure}

As seen in Fig.~\ref{fig:framework_and_controller}b), we perform this decoupling by leveraging the Configuration Space of the BR2, which is described in terms of strains: Curvature ($\kappa$) and Torsion ($\tau$).  It is then possible to create two maps: one between Actuation Space and Configuration Space and a second between Configuration Space and Task Space. The first map depends on the mechanical properties of the SCA, such as material, actuation, design, manufacturing, etc., and thus varies greatly - even with time.  Creating this map from physical principles is challenging and almost always relies on some data driven approach. The second map is purely kinematic and independent of the particulars of the BR2 manipulator. By creating an \textit{RL kinematic controller} that works in Configuration Space, i.e., the second map, we are able to ensure that the controller can be applied to different hardware configurations. To handle the first map between Actuation Space and Configuration Space, a \textit{local controller} is employed. It is used to iteratively refine the actuation through a correction loop (Fig.~\ref{fig:framework_and_controller}c). This allows the system to avoid the need for an custom Configuration-to-Actuation map and still achieve the desired configuration albeit iteratively. A general Configuration-to-Actuation map, created from data generated by a Cosserat rod model created for previous work ~\citep{ripperger2023}, is used for large movements and refined by the \textit{local controller}.

\subsection{RL Problem Formulation}
Following the MDP framework, we define the state space, action space, and reward as follows:
\paragraph{State space} The state space represents all possible states of the SCA. The state $s_t$ includes the current position $ p_t $ and orientation $ o_t $ of the end-effector, configuration parameters ($\kappa_t$ and $\tau_t$), the bounding box centroids of both the end-effector and the target from the base camera image $(\zeta_{be,t}, \zeta_{bg,t}$), the bounding box centroid of the target from the distal camera image ($\zeta_{dg,t}$), and a target visibility boolean based on the detection from the detection model ($v_t$). Formally, it can be represented as: $s_t = [p_t, o_t, \kappa_t, \tau_t, \zeta_{be,t}, \zeta_{bg,t}, \zeta_{dg,t}, v_t]$.

\paragraph{Action space} The action space encompasses all available actions. The actions consist of adjustments in curvature and torsion at each time step. Formally, they can be expressed as: $a_t = [\Delta \kappa, \Delta \tau]$. The actions are bounded between [-1, 1] and are scaled by a set of preset factors during stepping.

\paragraph{Reward} The reward function comprises the following components:\\
| \textit{The distance-based reward}, $r_d = e^{-\ln2 \left(40d/\pi\right)^2}$, encourages reduction in the Euclidean distance $d$ in meters between the current position of the end-effector and the target.\\
| \textit{The angle-based reward}, $r_a = e^{-\ln2 \left(8\alpha/\pi\right)^2}$, incentivizes alignment of the end-effector’s orientation with the target. Here, \( \alpha \) is the angle between the end-effector’s normal vector and the vector from the end-effector to the target.\\
| \textit{The visual information-based reward} \( r_i \), utilizes feedback from the distal camera to refine alignment with the target. This reward minimizes the visual discrepancy between the target's bounding box centroid and the center of the distal camera frame. It is given as $r_a = 5e^{-2\pi (d_i/l)^2} $ if the target is visible in distal camera, 0 otherwise. Here $d_i$ is the distance between the target's bounding box centroid in pixels, $\zeta_{dg,t}$, and the frame's center, $\zeta_c$. $l$ represents half the diagonal length of the distal camera frame, also in pixels.\\
| \textit{The task completion reward} $r_c$, reinforces successful completion of the task, which is assigned when the distance between the target's bounding box centroid and the distal camera's frame, $d_i$, is less than 100 pixels. The task completion reward is $r_c = 128$ if $d_i \leq 100$, 0 otherwise.\\
| \textit{The time penalty} $r_p = -10$ penalizes the agent for prolonged episode duration, encouraging efficient task completion. The total reward at each time step is given by $r_t = r_d + r_a + r_i + r_c + r_p$.

The reward function is designed to balance the contributions of each component while maintaining a hierarchy of priorities. Exponential terms create sharper gradients near desired values, providing stronger guidance as the agent approaches critical goals, such as minimizing Euclidean distance or achieving alignment. The scaling factors and specific values are chosen to prioritize key objectives. For instance, task completion carries the highest reward to emphasize goal achievement, while rewards such as the distance- and angle-based rewards, are scaled to encourage incremental progress without overshadowing the importance of overall success. This structure ensures that the agent focuses on the most critical aspects of the task, fostering effective learning.


\subsection{RL Training Framework}
Fig.~\ref{fig:framework_and_controller}a) illustrates the training framework for the \textit{RL kinematic controller} (black + blue paths). In this framework, Grounding DINO~\citep{liu2024groundingdinomarryingdino}, is used to detect bounding boxes from the images captured by the distal and base camera. From the base camera image, the positions of both the end-effector and the target are extracted, while from the distal camera image, only the position of the target is extracted when visible. These detections provide visual information that are processed to form part of the state input for the RL agent. Within this framework, the RL agent interacts with the environment iteratively. At each time step, the agent receives the current state of the SCA from the simulator and executes an action to adjust the SCA’s configuration. The simulator processes this control signal to update the state based on the underpinning model while the reward calculator evaluates the effectiveness of the agent's action in achieving the desired objectives. This feedback is then returned to the RL agent in the form of a scalar reward, allowing it to refine its control policy over iterative interactions with the environment.

\paragraph{Simulation environment setup} The simulation environment was built in Gazebo~\citep{1389727}, modeling the BR2 under the assumption of constant curvature and torsion. To represent the SCA, the simulation includes a series of 2 cm spheres spaced along the arm’s length, with cameras providing visual feedback from both the base and the distal tip. A 3 cm red sphere represents the target object in the workspace, which is enclosed on three sides to mirror real-world constraints. The environment was wrapped using Gymnasium~\citep{towers2024gymnasium}, enabling seamless integration with RL algorithms and standardized training pipelines.

\subsection{Evaluation Mertics}
To assess the performance of the trained RL agent, we define a set of evaluation metrics that capture both task success and control precision:
\paragraph{Success rate} The percentage of trials in which success is achieved, i.e., target centered in the distal camera image, or $d_i \leq 100$. Here $d_i$ is the distance in pixels between the target’s bounding box centroid and center of the distal camera’s frame.
\paragraph{Steps to goal} Defined as the number of iterations needed to achieve success.
\paragraph{Repeatability} Defined as the percentage of repeated outcomes on a point by point basis.

\subsection{Training and Evaluating the \textit{RL kinematic controller}}
\paragraph{Training} The training for the \textit{RL kinematic controller} is conducted entirely in simulation. Each training episode begins with the BR2 in a randomized initial configuration and the target placed at a random position within the workspace. An episode terminates either when the RL agent successfully completes the task, defined as centering the target in the distal camera frame (refer to success rate in the previous subsection), or when the agent exceeds the maximum allowed steps per episode, set to 8 in this study. The training process employs the implementation of the Soft Actor-Critic (SAC) algorithm~\citep{haarnoja2018soft} from Stable-Baselines3~\citep{stable-baselines3}, with a total training duration of 150k steps. Details on the hyperparameter settings and the reward plot during the training can be found in~\href{https://tinyurl.com/53f5vdje}{supplementary materials}.

\begin{figure}[h!]
\centering
\includegraphics[width=0.9\linewidth]{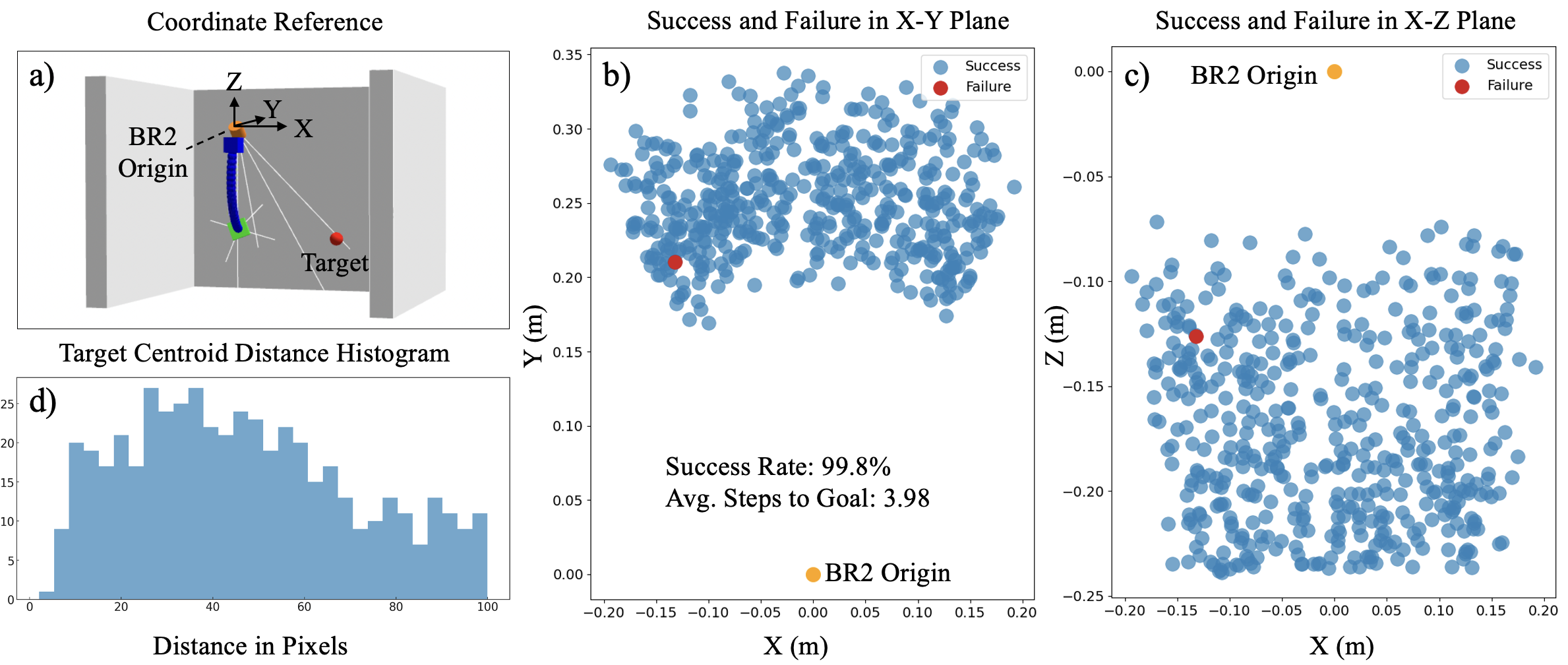}

\caption{\textbf{a)} Coordinate reference. \textbf{b)} and \textbf{c)} Scatter plots of the sampled target positions in the workspace. Each dot represents the outcome of an episode, where blue dots indicate successful goal-reaching and red dots represent failures. These plots illustrate the generalization capability of the trained RL kinematic controller across the workspace, with a high density of blue dots demonstrating robust performance. \textbf{d)} Histogram of the distance between the target bounding box centroid and the center of the distal camera frame.}
\label{fig:results_sim}

\end{figure}

\paragraph{Evaluation} The trained \textit{RL kinematic controller} is evaluated in simulation to ensure its performance in visual servoing tasks before the sim-to-real transfer. The evaluation consisted of 500 randomized episodes, with the testing procedures mirroring the training, including randomized initial configurations for the BR2 and target, and the same termination criteria. The sampled target positions were distributed throughout the workspace, as shown in Fig.~\ref{fig:results_sim} b) and c), ensuring that the evaluation comprehensively covered a wide range of configurations. 

The evaluation focused on key metrics: success rate, average steps to task completion, and the target centroid distance from the center of the distal camera image. The trained RL kinematic controller achieved a success rate of 99.8\%, requiring 3.98 steps on average to complete the visual servoing task. As shown in Fig.~\ref{fig:results_sim}b) and c), the scatter plots of target positions represent the outcomes of each episode, where blue dots indicate successful episodes carried out by the trained policy, and red dots represent failures. The high density of blue dots demonstrates the controller's robust generalization across the workspace.


\subsection{Deployment Framework}
The deployment framework, also illustrated in Fig.~\ref{fig:framework_and_controller}a), integrates the trained \textit{RL kinematic controller} with the \textit{local controller} (black + dashed paths). The process begins with an input containing target end-effector pose information, provided as prompts to the object detector along with images from the distal and base cameras.  The RL kinematic controller processes this input and other kinematic data to determine the goal configuration that aligns the SCA with the specified goal. Once the target configuration is established, it is passed to the \textit{local controller}, which translates this kinematic goal into actuation.

\paragraph{\textit{Local controller}} The \textit{local controller}, shown in Fig.~\ref{fig:framework_and_controller}c), is used to generate actuations to achieve the desired configuration using an iterative three step process:  1) Using the tip sensor pose data, the current arm configuration is estimated.  2) The error between the estimated and desired configuration is used by a heuristic (see~\href{https://tinyurl.com/53f5vdje}{supplementary materials} for more details) to produce a change in actuation. 3) The new actuation is applied and after reaching steady state, it returns to Step 1. This is iterated until the desired accuracy is achieved. This method avoids being over reliant on a custom Configuration-to-Actuation map unique to each BR2. To set the initial configuration a general but likely inaccurate Configuration-to-Actuation map is used and the above configuration corrective loop is used to achieve the goal configuration.

\section{Sim-to-Real Transfer Experiments}


\subsection{Hardware Setup}
Our hardware setup closely mirrors the simulation environment, featuring the BR2 equipped with two cameras | a base camera to capture a fixed view of the workspace and a distal camera mounted on the distal tip. The setup includes target objects positioned on stands of varying heights, as well as a tracking instrument (Polhemus Patriot) to monitor the pose of the end-effector tip. The base camera is angled 45\degree{} downward to provide a comprehensive view of the workspace, while the distal camera is aligned with the axial line of the arm, offering a direct view of the tip’s orientation relative to the target. The \textit{local controller} is implemented as a simple closed-loop controller to achieve the desired RL policy configurations without the need for a custom Configuration-to-Actuation map.



\subsection{Testing Procedure}
To validate the effectiveness of the deployed RL policy, we designed a testing proedure to challenge the BR2 to reach various target positions within the workspace, which is as follows:\\
| \textit{Initial setup}: Each test starts with the BR2 in a random configuration with the tip visible in the base camera and the target object placed at one of 50 test positions through out the workspace and within the frame of the base camera.\\
| \textit{Planning and execution}: The RL policy plans for the next configuration and the \textit{local controller} executes the plan. The control loop continues until task completion or time out.\\
| \textit{Evaluation, repositioning and repeat}: After each test, the target object is repositioned to a new location. For each new target position, the procedure is repeated, with key metrics—such as positioning accuracy, alignment with the target, steps taken and success rate recorded for each trial.\\
| \textit{Additional weights}: To evaluate the effectiveness of the local controller to overcome variations in the Configuration-to-Actuation map (Fig.~\ref{fig:framework_and_controller}b), weights (10g, 15g and 20g) were added to the tip of the BR2.  A subset of the test points were then tested in the same manner noted above.\\

\subsection{Results and Discussion}
\begin{figure}[h]

\centering
\includegraphics[width=\linewidth]{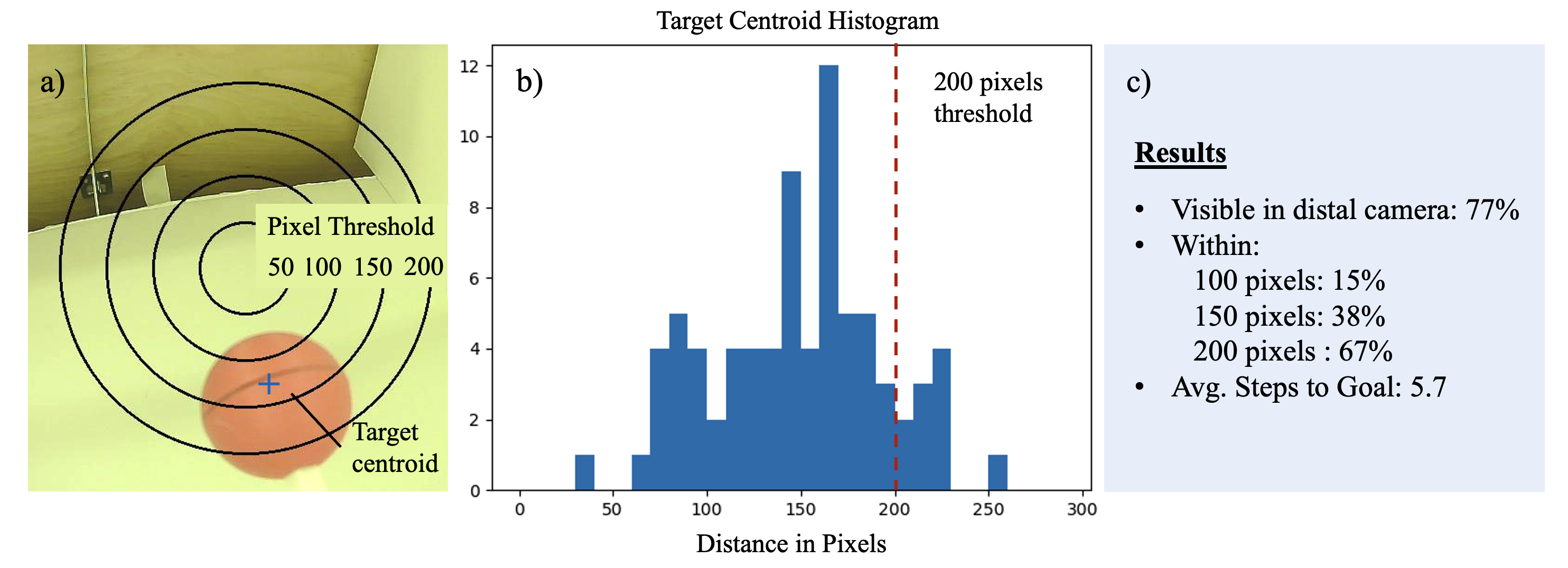}

\caption{Tip View Results. \textbf{a)} A representative view from the tip camera.  The rings indicate pixel distance thresholds. \textbf{b)} A histogram of best centroid distances in pixels.  \textbf{c)} A summary of the results of testing.}
\label{fig:results_1}

\end{figure}

\begin{figure}[h]
\centering
\includegraphics[width=0.9\linewidth]{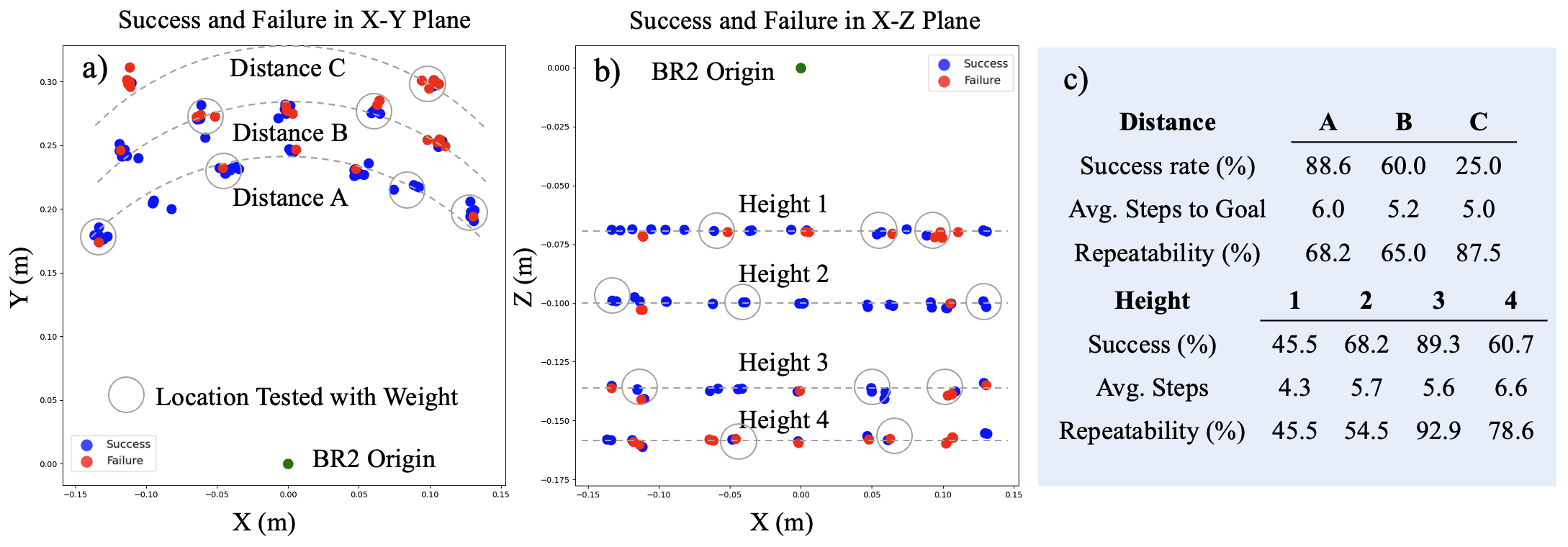}
\caption{Regional Results. \textbf{a)} A top down view of the test area showing three distances tested. \textbf{b)} A view into the test area showing the four test heights. Locations of tip weight testing are marked with circles. \textbf{c)} A summary of the results based on test point region.}
\label{fig:results_region}

\end{figure}

Three levels of success were analyzed based on the target centroid's distance from the distal camera center measured: 100 pixels, 150 pixels and 200 pixels (Fig.~\ref{fig:results_1}a).  Fifty target positions throughout the workspace were tested twice for a total of 100 samples.  Overall it was detected in the tip view by Grounding DINO in 77$\%$ of the tests.  The histogram in Fig.~\ref{fig:results_1}b shows that few targets centroids were brought within the 100 pixel threshold used in the simulation.  This is due to challenges in performing small corrective movements on the real hardware as these corrections were close to the controller error threshold.  Fig.~\ref{fig:results_1}c demonstrates the results for the different levels of success.  These results are weaker than the results observed in the simulation, but this is to be expected given the challenges of sim-to-real transfer.  We selected a 200 pixel threshold as the standard for the hardware tests because, at this threshold, it is still possible to fully view the target in the tip camera and the image is useful for tasks such as inspection and counting.  

A regional analysis of success and failure can be seen in Fig.~\ref{fig:results_region}, but still demonstrate that policy works on hardware.  This demonstrates the policy is most accurate in the central region, middle height and close to the reach of the BR2.  This region uses the least rotational actuation which leads to less model error and there is less perspective distortion with the world image.  It is worth noting that in every test case, even if it was not able to view the target with the tip camera, the BR2 arm servoed toward the target. In addition to success, we examined the repeatability of the results.  There was an overall repeatability of 70$\%$.  A regional breakdown can be seen in Fig.~\ref{fig:results_region}c.  Naturally strong success or failure leads to repeatability.  The results indicate that the lower test points in addition to having lower success, are also inconsistent in their performance. The primary failure mode was excessive curvature, which appears to be driven by differences in real world BR2 response compared to the constant curvature and torsion assumption used in training.  Additionally, a lack of depth information likely contributes to failures with greater target distance.

Weights (10g, 15g and 20g) were attached to the tip of the BR2 and a subset of the test points were tested and for the 200 pixel threshold.  The 10g achieved 57.1$\%$ and the 15g and 20g both achieved 50$\%$.  These results indicate that the controller can overcome map inaccuracies.  However, as the weight increased, the system struggled to achieve the more extreme positions as actuations reached the pressure limits. While a non-negligible gap remains, our results demonstrate meaningful sim-to-real transfer under realistic, minimally-instrumented conditions. Unlike prior works relying on extensive sensing or hardware-specific tuning, our single policy—trained entirely in simulation—handles randomized 3D visual servoing using only visual feedback. This highlights the robustness and generality of our approach for SCAs. Please refer to~\href{https://tinyurl.com/53f5vdje}{supplementary materials} for detailed discussion and more analysis of the sim-to-real errors.

\section{Conclusion and Future Work}
In this work, we demonstrated a zero shot sim-to-real transfer of an RL policy to visually servo a BR2.  The policy was trained on a reduced order constant curvature and torsion model based simulation and successfully tested on real hardware. In future work, we hope to expand on the success of our present work. We aim to improve the system's success rate by refining learning and control strategies. Particularly, we plan to enhance tip view alignment for tighter centering thresholds and expand the workspace by adding degrees of freedom, enabling tasks beyond image centering, such as grasping or multi-angle inspection. Lastly, we also intend to leverage to the power of Grounding DINO to work with a variety of targets in unstructured environments.


\section*{Ackowledgements}{This work is funded from a joint NSF-USDA COALESCE grant 2021-67021-34418 and NSF CPS Frontier \#1954556.  Additional funding by AIFARMS National AI Institute in Agriculture, backed by Agriculture and Food Research Initiative (Grant Number: 2020-67021-32799).}

\newpage
\section*{Supplementary Materials}
\subsection*{Related Works}

While SCAs exhibit promising adaptivity and safety in various applications, the inherent complexity of their control remains a significant challenge(controlling SCAs is particularly challenging due to their infinite degrees of freedom and inherent limitations in available methods for sensing their deformed shapes ~\citep{rus2015design}. To address this, various control strategies have been developed, each with unique strengths and limitations. In the following section, we review these strategies, highlighting the evolution from classical model-based controls to more flexible learning-based approaches, which pave the way for our model-free solution.

\paragraph{SCA Control}
Control strategies for SCAs can generally be categorized into classical model-based methods, model-based learning approaches, and model-free learning methods. Classical model-based controls typically rely on analytical formulations for estimating kinematics~\citep{webster2010design}, with constant curvature assumptions often forming the basis for these models. While these approaches simplify the control problem, they struggle with non-linear deformations and dynamic uncertainties in real-world scenarios. Advanced techniques, such as Cosserat rod models, enhance accuracy by capturing continuous deformations but require significant expertise to implement and are computationally intensive. This complexity, combined with the need for precise and cost-effective spatial feedback, restricts their widespread adoption within the field~\citep{satheeshbabu2020continuous}.
Similarly, learning-based methods have explored combining physical principles with data-driven modeling to improve flexibility and control accuracy~\citep{falotico2024learning}. However, challenges like computational complexity and sim-to-real transfer still persist, especially for highly deformable systems like SCAs.

To address these limitations, model-based learning approaches have been introduced, leveraging data-driven models such as Artificial Neural Networks (ANNs) and Gaussian Processes (GPs) to improve control accuracy. These methods attempt to bypass some of the modeling assumptions in classical approaches, yet they still face challenges, such as managing workspace discontinuities and ensuring stable inverse kinematics under varying loads~\citep{d2001learning}. Notable examples include a novel RL method based on the Cosserat rod model, where fuzzy reinforcement learning (FRL) was applied along with optimization techniques like the Taguchi method and genetic algorithms (GA). This approach enabled the continuum robot to perform stable and accurate trajectory tracking~\citep{goharimanesh2020fuzzy}. Another study applied GP-based RL to the constant curvature model, addressing forward kinematics problems~\citep{thuruthel2018model}. Deep Q-learning has also been explored for position control in cable-driven soft manipulators~\citep{wu2020position}. While these model-based learning approaches offer increased flexibility, they have not been extensively applied to complex SCAs with spatial bending and rotational capabilities, such as the BR2, which possess enhanced workspace and task versatility.

More recently, model-free RL has gained traction as a robust solution for SCA control, especially in cases where modeling complexities or environmental variabilities pose challenges to model-based methods~\citep{you2017model, zhang2017toward}. Early studies employed Q-learning for static position control of SCAs, but were often restricted to planar tasks with discretized actions. Advanced methods, such as Deep Deterministic Policy Gradient (DDPG), have shown promise in enabling continuous control over larger state-action spaces, which aligns with the complex, adaptive behaviors required for SCAs in real-world applications~\citep{satheeshbabu2020continuous}.

While these methods illustrate significant advancements, they highlight the trade-offs between sensing complexity and control performance. They either simplify sensing at the expense of task space complexity(~\citep{wu2020position, satheeshbabu2019open, goharimanesh2020fuzzy, morimoto2021model}) or rely on extensive sensory setups to improve robustness, such as the multi-sensor approach of ~\citep{thuruthel2018model} or the Vicon-based strategy of ~\citep{satheeshbabu2020continuous}.
\paragraph{Visual Servoing with SCAs}
Visual servoing has also emerged as a viable strategy for SCAs, particularly due to the challenges in obtaining accurate model-based pose control.
VS approaches for SCAs leverage visual feedback for tasks like pose control and adaptive tracking but often rely on structured setups and detailed sensing systems. Early studies, such as ~\citep{xu2019underwater, xu2021visual}, used fixed cameras (eye-to-hand), while ~\citep{wang2013visual, wang2016visual} introduced eye-in-hand configurations with adaptive controllers for constrained environments. More recent works, like ~\citep{wang2020eye}, combined monocular feedback with strain sensors for 2D tasks. Additionally, advancements in eye-in-hand, image-based visual servoing within three-dimensional spaces have incorporated neural networks for robust feature extraction, addressing discrepancies between predicted and actual actuation states by visually comparing real-time images against target images~\citep{kamtikar2022visual}. ~\citep{albeladi2022hybrid} proposed a hybrid VS approach, combining eye-in-hand and eye-to-hand configurations to enhance tracking adaptability across diverse workspaces. Similarly, ~\citep{norouzi2021switching} introduced a switching image-based visual servoing (IBVS) method for continuum robots, emphasizing precision and error-handling to improve stability in constrained surgical environments.
Despite their advancements, these methods are predominantly model-based and rely on precise sensing setups, limiting scalability in dynamic, unstructured settings. Recently, \citep{liu2020real} employed model-free reinforcement learning for visual servoing of soft continuum arms. However, their approach focused on a 2D task space and did not address zero-shot sim-to-real transfer as all training and testing were conducted on physical hardware. These constraints highlight the persistent challenges in adapting RL-based methods for visual servoing in more complex, dynamic, and unstructured environments with simplified sensing systems that rely on minimal sensory inputs. To bridge these gaps, our work introduces a minimum-sensing RL framework designed to achieve robust zero-shot sim-to-real transfer for 3D navigation and visual servoing tasks.

\subsection*{Sim-to-real Performance Discrepancy}
Direct comparisons of sim-to-real performance across existing literature are inherently difficult, as prior works vary widely in task formulation, success metrics, sensing modalities, and evaluation criteria. Some focus on static target positioning~\citep{satheeshbabu2019open}, others on trajectory tracking~\citep{satheeshbabu2020continuous, goharimanesh2020fuzzy}, and still others train and evaluate entirely on physical hardware~\citep{morimoto2021model, liu2020real}. These differences make it challenging to define a common baseline for success or degradation.

Despite these variations, our results demonstrate meaningful transfer. Our policy, trained entirely in simulation, achieves a 67\% success rate in hardware deployment with no fine-tuning, using only visual inputs from a minimal two-camera setup and no internal sensing. In contrast to systems that rely on motion capture~\citep{thuruthel2018model, satheeshbabu2020continuous} or custom instrumentation, our approach performs closed-loop visual servoing in 3D using only monocular bounding box cues.

Moreover, many existing methods evaluate success under structured or repeated scenarios (e.g., fixed goals, known workspaces, constrained actuation), whereas our evaluation involves randomized 3D targets and goal-invariant control. While the sim-to-real performance gap (99.8\% to 67\%) is non-negligible, it is a reasonable outcome given the reduced sensing and increased variability in our task setup. Overall, our results highlight the feasibility of sim-to-real transfer for model-free visual servoing under realistic constraints, setting a precedent for future work that prioritizes generality and scalability over lab-specific accuracy.

\begin{figure}[h!]
\centering
\includegraphics[width=0.9\linewidth]{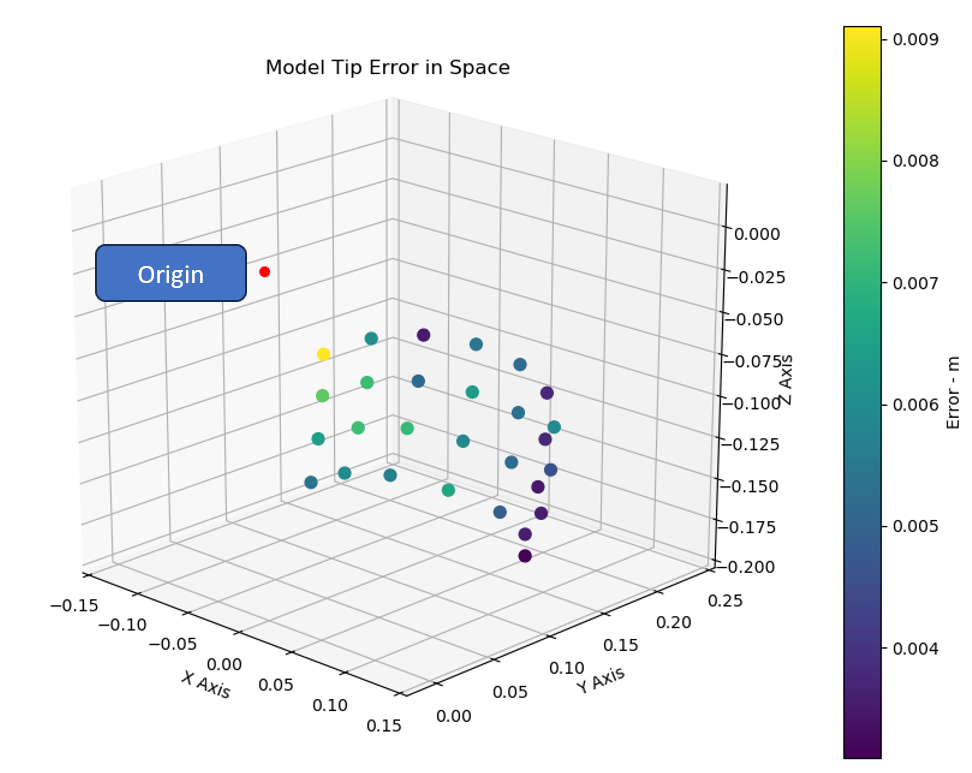}

\caption{Tip Error In Space - This graph shows the Euclidean Distance error between the hardware tip position and model tip position for the same configuration. Data was collected throughout the workspace.} 
\label{fig:model_error}
\end{figure}

\subsection*{Model versus Real World Hardware Error Comparison}
To understand and quantify the error between the model and hardware, tip position data was collected on the hardware and compared to that of the model.  To achieve this, a set of curvature and torsion goals were selected to span the workspace.  The hardware controller was then used to achieve each of these configuration goals and the end position was recorded.  The final configuration achieved (i.e. the estimated configuration) was then used in the model to find the tip position.  The Euclidean distance error between the model tip position and the hardware tip position can be seen in Figure \ref{fig:model_error}.  The average error is 5.5mm and all measured points are less than 1cm error.  There is an asymmetry in the results which is likely a results of the construction of the BR2 and variations in the behavior of the two rotational actuators.  The error is relatively even and low throughout the workspace. Despite the left-right split, there is not a significant difference in failures between the two sides (9 on the left and 11 on the right). There is some evidence of more extreme positions having more error, but this is masked somewhat by the asymmetry.  Higher torsions typically result in the greatest error between the model and real world as the real torsion does not match the constant torsion model well as torsion increases.

\bibliography{ref}

\begin{thebibliography}{40}
\providecommand{\natexlab}[1]{#1}
\providecommand{\url}[1]{\texttt{#1}}
\expandafter\ifx\csname urlstyle\endcsname\relax
  \providecommand{\doi}[1]{doi: #1}\else
  \providecommand{\doi}{doi: \begingroup \urlstyle{rm}\Url}\fi

\bibitem[AlBeladi et~al.(2022)AlBeladi, Ripperger, Hutchinson, and Krishnan]{albeladi2022hybrid}
Ali AlBeladi, Evan Ripperger, Seth Hutchinson, and Girish Krishnan.
\newblock Hybrid eye-in-hand/eye-to-hand image based visual servoing for soft continuum arms.
\newblock \emph{IEEE Robotics and Automation Letters}, 7\penalty0 (4):\penalty0 11298--11305, 2022.

\bibitem[Banerjee et~al.(2018)Banerjee, Tse, and Ren]{banerjee2018soft}
Hritwick Banerjee, Zion Tsz~Ho Tse, and Hongliang Ren.
\newblock Soft robotics with compliance and adaptation for biomedical applications and forthcoming challenges.
\newblock \emph{Int. J. Robot. Autom}, 33\penalty0 (1):\penalty0 68--80, 2018.

\bibitem[Chen et~al.(2021)Chen, Pang, Cao, Tan, and Cao]{chen2021soft}
Shoue Chen, Yaokun Pang, Yunteng Cao, Xiaobo Tan, and Changyong Cao.
\newblock Soft robotic manipulation system capable of stiffness variation and dexterous operation for safe human--machine interactions.
\newblock \emph{Advanced Materials Technologies}, 6\penalty0 (5):\penalty0 2100084, 2021.

\bibitem[Chen et~al.(2022)Chen, Zhang, Huang, Cao, and Liu]{chen2022review}
Xiaoqian Chen, Xiang Zhang, Yiyong Huang, Lu~Cao, and Jinguo Liu.
\newblock A review of soft manipulator research, applications, and opportunities.
\newblock \emph{Journal of Field Robotics}, 39\penalty0 (3):\penalty0 281--311, 2022.

\bibitem[D'Souza et~al.(2001)D'Souza, Vijayakumar, and Schaal]{d2001learning}
Aaron D'Souza, Sethu Vijayakumar, and Stefan Schaal.
\newblock Learning inverse kinematics.
\newblock In \emph{Proceedings 2001 IEEE/RSJ International Conference on Intelligent Robots and Systems. Expanding the Societal Role of Robotics in the the Next Millennium (Cat. No. 01CH37180)}, volume~1, pages 298--303. IEEE, 2001.

\bibitem[Falotico et~al.(2024)Falotico, Donato, Alessi, Setti, Nazeer, Agabiti, Caradonna, Bianchi, Piqu{\'e}, Ansari, et~al.]{falotico2024learning}
Egidio Falotico, Enrico Donato, Carlo Alessi, Elisa Setti, Muhammad~Sunny Nazeer, Camilla Agabiti, Daniele Caradonna, Diego Bianchi, Francesco Piqu{\'e}, Yasmin~Tauqeer Ansari, et~al.
\newblock Learning controllers for continuum soft manipulators: Impact of modeling and looming challenges.
\newblock \emph{Advanced Intelligent Systems}, page 2400344, 2024.

\bibitem[Goharimanesh et~al.(2020)Goharimanesh, Mehrkish, and Janabi-Sharifi]{goharimanesh2020fuzzy}
Masoud Goharimanesh, Ali Mehrkish, and Farrokh Janabi-Sharifi.
\newblock A fuzzy reinforcement learning approach for continuum robot control.
\newblock \emph{Journal of Intelligent \& Robotic Systems}, 100\penalty0 (3):\penalty0 809--826, 2020.

\bibitem[Haarnoja et~al.(2018)Haarnoja, Zhou, Hartikainen, Tucker, Ha, Tan, Kumar, Zhu, Gupta, Abbeel, et~al.]{haarnoja2018soft}
Tuomas Haarnoja, Aurick Zhou, Kristian Hartikainen, George Tucker, Sehoon Ha, Jie Tan, Vikash Kumar, Henry Zhu, Abhishek Gupta, Pieter Abbeel, et~al.
\newblock Soft actor-critic algorithms and applications.
\newblock \emph{arXiv preprint arXiv:1812.05905}, 2018.

\bibitem[Janabi-Sharifi et~al.(2021)Janabi-Sharifi, Jalali, and Walker]{9420666}
Farrokh Janabi-Sharifi, Amir Jalali, and Ian~D. Walker.
\newblock Cosserat rod-based dynamic modeling of tendon-driven continuum robots: A tutorial.
\newblock \emph{IEEE Access}, 9:\penalty0 68703--68719, 2021.
\newblock \doi{10.1109/ACCESS.2021.3077186}.

\bibitem[Kamtikar et~al.(2022)Kamtikar, Marri, Walt, Uppalapati, Krishnan, and Chowdhary]{kamtikar2022visual}
Shivani Kamtikar, Samhita Marri, Benjamin Walt, Naveen~Kumar Uppalapati, Girish Krishnan, and Girish Chowdhary.
\newblock Visual servoing for pose control of soft continuum arm in a structured environment.
\newblock \emph{IEEE Robotics and Automation Letters}, 7\penalty0 (2):\penalty0 5504--5511, 2022.

\bibitem[Koenig and Howard(2004)]{1389727}
Nathan~P. Koenig and Andrew Howard.
\newblock Design and use paradigms for gazebo, an open-source multi-robot simulator.
\newblock In \emph{2004 IEEE/RSJ International Conference on Intelligent Robots and Systems (IROS) (IEEE Cat. No.04CH37566)}, volume~3, pages 2149--2154 vol.3, 2004.
\newblock \doi{10.1109/IROS.2004.1389727}.

\bibitem[Leguizamo et~al.(2022)Leguizamo, Yang, Lee, and Sarkar]{leguizamo2022deep}
David~Felipe Leguizamo, Hsin-Jung Yang, Xian~Yeow Lee, and Soumik Sarkar.
\newblock Deep reinforcement learning for robotic control with multi-fidelity models.
\newblock \emph{IFAC-PapersOnLine}, 55\penalty0 (37):\penalty0 193--198, 2022.

\bibitem[Li et~al.(2024)Li, Ma, Hu, Zhang, Liu, and Sun]{li2024vision}
Jinzhou Li, Jie Ma, Yujie Hu, Li~Zhang, Zhijie Liu, and Shiying Sun.
\newblock Vision-based reinforcement learning control of soft robot manipulators.
\newblock \emph{Robotic Intelligence and Automation}, 44\penalty0 (6):\penalty0 783--790, 2024.

\bibitem[Liu et~al.(2020)Liu, Cai, Lu, Wang, and Wang]{liu2020real}
Naijun Liu, Yinghao Cai, Tao Lu, Rui Wang, and Shuo Wang.
\newblock Real--sim--real transfer for real-world robot control policy learning with deep reinforcement learning.
\newblock \emph{Applied Sciences}, 10\penalty0 (5):\penalty0 1555, 2020.

\bibitem[Liu et~al.(2024)Liu, Zeng, Ren, Li, Zhang, Yang, Jiang, Li, Yang, Su, Zhu, and Zhang]{liu2024groundingdinomarryingdino}
Shilong Liu, Zhaoyang Zeng, Tianhe Ren, Feng Li, Hao Zhang, Jie Yang, Qing Jiang, Chunyuan Li, Jianwei Yang, Hang Su, Jun Zhu, and Lei Zhang.
\newblock Grounding dino: Marrying dino with grounded pre-training for open-set object detection, 2024.
\newblock URL \url{https://arxiv.org/abs/2303.05499}.

\bibitem[Mnih et~al.(2013)Mnih, Kavukcuoglu, Silver, Graves, Antonoglou, Wierstra, and Riedmiller]{mnih2013playing}
Volodymyr Mnih, Koray Kavukcuoglu, David Silver, Alex Graves, Ioannis Antonoglou, Daan Wierstra, and Martin Riedmiller.
\newblock Playing atari with deep reinforcement learning, 2013.

\bibitem[Morimoto et~al.(2021)Morimoto, Nishikawa, Niiyama, and Kuniyoshi]{morimoto2021model}
Ryota Morimoto, Satoshi Nishikawa, Ryuma Niiyama, and Yasuo Kuniyoshi.
\newblock Model-free reinforcement learning with ensemble for a soft continuum robot arm.
\newblock In \emph{2021 IEEE 4th International Conference on Soft Robotics (RoboSoft)}, pages 141--148. IEEE, 2021.

\bibitem[Norouzi-Ghazbi and Janabi-Sharifi(2021)]{norouzi2021switching}
Somayeh Norouzi-Ghazbi and Farrokh Janabi-Sharifi.
\newblock A switching image-based visual servoing method for cooperative continuum robots.
\newblock \emph{Journal of Intelligent \& Robotic Systems}, 103\penalty0 (3):\penalty0 42, 2021.

\bibitem[Raffin et~al.(2021)Raffin, Hill, Gleave, Kanervisto, Ernestus, and Dormann]{stable-baselines3}
Antonin Raffin, Ashley Hill, Adam Gleave, Anssi Kanervisto, Maximilian Ernestus, and Noah Dormann.
\newblock Stable-baselines3: Reliable reinforcement learning implementations.
\newblock \emph{Journal of Machine Learning Research}, 22\penalty0 (268):\penalty0 1--8, 2021.
\newblock URL \url{http://jmlr.org/papers/v22/20-1364.html}.

\bibitem[Ripperger and Krishnan(2023)]{ripperger2023}
Evan Ripperger and Girish Krishnan.
\newblock Design space enumerations for pneumatically actuated soft continuum manipulators.
\newblock In \emph{Volume 8: 47th Mechanisms and Robotics Conference (MR)}, page V008T08A097, 08 2023.
\newblock \doi{10.1115/DETC2023-116930}.
\newblock URL \url{https://doi.org/10.1115/DETC2023-116930}.

\bibitem[Rus and Tolley(2015)]{rus2015design}
Daniela Rus and Michael~T Tolley.
\newblock Design, fabrication and control of soft robots.
\newblock \emph{Nature}, 521\penalty0 (7553):\penalty0 467--475, 2015.

\bibitem[Satheeshbabu et~al.(2019)Satheeshbabu, Uppalapati, Chowdhary, and Krishnan]{satheeshbabu2019open}
Sreeshankar Satheeshbabu, Naveen~Kumar Uppalapati, Girish Chowdhary, and Girish Krishnan.
\newblock Open loop position control of soft continuum arm using deep reinforcement learning.
\newblock In \emph{2019 International Conference on Robotics and Automation (ICRA)}, pages 5133--5139. IEEE, 2019.

\bibitem[Satheeshbabu et~al.(2020)Satheeshbabu, Uppalapati, Fu, and Krishnan]{satheeshbabu2020continuous}
Sreeshankar Satheeshbabu, Naveen~K Uppalapati, Tianshi Fu, and Girish Krishnan.
\newblock Continuous control of a soft continuum arm using deep reinforcement learning.
\newblock In \emph{2020 3rd IEEE International Conference on Soft Robotics (RoboSoft)}, pages 497--503. IEEE, 2020.

\bibitem[Schulman et~al.(2017)Schulman, Wolski, Dhariwal, Radford, and Klimov]{schulman2017proximal}
John Schulman, Filip Wolski, Prafulla Dhariwal, Alec Radford, and Oleg Klimov.
\newblock Proximal policy optimization algorithms.
\newblock \emph{arXiv preprint arXiv:1707.06347}, 2017.

\bibitem[Thuruthel et~al.(2018)Thuruthel, Falotico, Renda, and Laschi]{thuruthel2018model}
Thomas~George Thuruthel, Egidio Falotico, Federico Renda, and Cecilia Laschi.
\newblock Model-based reinforcement learning for closed-loop dynamic control of soft robotic manipulators.
\newblock \emph{IEEE Transactions on Robotics}, 35\penalty0 (1):\penalty0 124--134, 2018.

\bibitem[Till et~al.(2019)Till, Aloi, and Rucker]{till2019real}
John Till, Vincent Aloi, and Caleb Rucker.
\newblock Real-time dynamics of soft and continuum robots based on cosserat rod models.
\newblock \emph{The International Journal of Robotics Research}, 38\penalty0 (6):\penalty0 723--746, 2019.

\bibitem[Towers et~al.(2024)Towers, Kwiatkowski, Terry, Balis, De~Cola, Deleu, Goul{\~a}o, Kallinteris, Krimmel, KG, et~al.]{towers2024gymnasium}
Mark Towers, Ariel Kwiatkowski, Jordan Terry, John~U Balis, Gianluca De~Cola, Tristan Deleu, Manuel Goul{\~a}o, Andreas Kallinteris, Markus Krimmel, Arjun KG, et~al.
\newblock Gymnasium: A standard interface for reinforcement learning environments.
\newblock \emph{arXiv preprint arXiv:2407.17032}, 2024.

\bibitem[Uppalapati and Krishnan(2021)]{uppalapati2018design}
Naveen~Kumar Uppalapati and Girish Krishnan.
\newblock Design and modeling of soft continuum manipulators using parallel asymmetric combination of fiber-reinforced elastomers.
\newblock \emph{Journal of Mechanisms and Robotics}, 13, 2 2021.
\newblock ISSN 1942-4302.
\newblock \doi{10.1115/1.4048223}.

\bibitem[Venter and Dirven(2017)]{venter2017self}
Dean Venter and Steven Dirven.
\newblock Self morphing soft-robotic gripper for handling and manipulation of delicate produce in horticultural applications.
\newblock In \emph{2017 24th International Conference on Mechatronics and Machine Vision in Practice (M2VIP)}, pages 1--6. IEEE, 2017.

\bibitem[Wang et~al.(2013)Wang, Chen, Yu, Deng, Wang, and Pfeifer]{wang2013visual}
Hesheng Wang, Weidong Chen, Xiaojin Yu, Tao Deng, Xiaozhou Wang, and Rolf Pfeifer.
\newblock Visual servo control of cable-driven soft robotic manipulator.
\newblock In \emph{2013 IEEE/RSJ International Conference on Intelligent Robots and Systems}, pages 57--62. IEEE, 2013.

\bibitem[Wang et~al.(2016)Wang, Yang, Liu, Chen, Liang, and Pfeifer]{wang2016visual}
Hesheng Wang, Bohan Yang, Yuting Liu, Weidong Chen, Xinwu Liang, and Rolf Pfeifer.
\newblock Visual servoing of soft robot manipulator in constrained environments with an adaptive controller.
\newblock \emph{IEEE/ASME transactions on mechatronics}, 22\penalty0 (1):\penalty0 41--50, 2016.

\bibitem[Wang et~al.(2020)Wang, Fang, Wang, Xie, Lee, Ho, Tang, Lam, and Kwok]{wang2020eye}
Xiaomei Wang, Ge~Fang, Kui Wang, Xiaochen Xie, Kit-Hang Lee, Justin~DL Ho, Wai~Lun Tang, James Lam, and Ka-Wai Kwok.
\newblock Eye-in-hand visual servoing enhanced with sparse strain measurement for soft continuum robots.
\newblock \emph{IEEE Robotics and Automation Letters}, 5\penalty0 (2):\penalty0 2161--2168, 2020.

\bibitem[Webster~III and Jones(2010)]{webster2010design}
Robert~J Webster~III and Bryan~A Jones.
\newblock Design and kinematic modeling of constant curvature continuum robots: A review.
\newblock \emph{The International Journal of Robotics Research}, 29\penalty0 (13):\penalty0 1661--1683, 2010.

\bibitem[Wu et~al.(2020)Wu, Gu, Li, Zhang, Chepinskiy, Wang, Zhilenkov, Krasnov, and Chernyi]{wu2020position}
Qiuxuan Wu, Yueqin Gu, Yancheng Li, Botao Zhang, Sergey~A Chepinskiy, Jian Wang, Anton~A Zhilenkov, Aleksandr~Y Krasnov, and Sergei Chernyi.
\newblock Position control of cable-driven robotic soft arm based on deep reinforcement learning.
\newblock \emph{Information}, 11\penalty0 (6):\penalty0 310, 2020.

\bibitem[Xu et~al.(2019)Xu, Wang, Wang, Au, and Chen]{xu2019underwater}
Fan Xu, Hesheng Wang, Jingchuan Wang, Kwok Wai~Samuel Au, and Weidong Chen.
\newblock Underwater dynamic visual servoing for a soft robot arm with online distortion correction.
\newblock \emph{IEEE/ASME Transactions on Mechatronics}, 24\penalty0 (3):\penalty0 979--989, 2019.

\bibitem[Xu et~al.(2021)Xu, Wang, Chen, and Miao]{xu2021visual}
Fan Xu, Hesheng Wang, Weidong Chen, and Yanzi Miao.
\newblock Visual servoing of a cable-driven soft robot manipulator with shape feature.
\newblock \emph{IEEE Robotics and Automation Letters}, 6\penalty0 (3):\penalty0 4281--4288, 2021.

\bibitem[Xun et~al.(2023)Xun, Zheng, and Kruszewski]{xun2023cosseratrodbaseddynamicmodeling}
Lingxiao Xun, Gang Zheng, and Alexandre Kruszewski.
\newblock Cosserat-rod based dynamic modeling of soft slender robot interacting with environment, 2023.
\newblock URL \url{https://arxiv.org/abs/2307.06261}.

\bibitem[You et~al.(2017)You, Zhang, Chen, Liu, Wang, Jiang, and Chen]{you2017model}
Xuanke You, Yixiao Zhang, Xiaotong Chen, Xinghua Liu, Zhanchi Wang, Hao Jiang, and Xiaoping Chen.
\newblock Model-free control for soft manipulators based on reinforcement learning.
\newblock In \emph{2017 IEEE/RSJ international conference on intelligent robots and systems (IROS)}, pages 2909--2915. IEEE, 2017.

\bibitem[Zhang et~al.(2017)Zhang, Cao, Zilberstein, Wu, and Chen]{zhang2017toward}
Haochong Zhang, Rongyun Cao, Shlomo Zilberstein, Feng Wu, and Xiaoping Chen.
\newblock Toward effective soft robot control via reinforcement learning.
\newblock In \emph{Intelligent Robotics and Applications: 10th International Conference, ICIRA 2017, Wuhan, China, August 16--18, 2017, Proceedings, Part I 10}, pages 173--184. Springer, 2017.

\bibitem[Zongxing et~al.(2020)Zongxing, Wanxin, and Liping]{zongxing2020research}
Lu~Zongxing, Li~Wanxin, and Zhang Liping.
\newblock Research development of soft manipulator: A review.
\newblock \emph{Advances in Mechanical Engineering}, 12\penalty0 (8):\penalty0 1687814020950094, 2020.

\end{thebibliography}

\end{document}